\documentclass[10pt,twocolumn,letterpaper]{article}

\usepackage[pagenumbers]{cvpr} 

\usepackage{colortbl}
\definecolor{cvprblue}{rgb}{0.21,0.49,0.74}
\usepackage[pagebackref,breaklinks,colorlinks,allcolors=cvprblue]{hyperref}

\def\paperID{15} 
\def\confName{CVPR}
\def\confYear{2025}

\title{Visual Variational Autoencoder Prompt Tuning}

\author{
\textbf{Xi Xiao}\textsuperscript{1}\thanks{Equal contribution} \quad
\textbf{Yunbei Zhang}\textsuperscript{2}\footnotemark[1] \quad
\textbf{Yanshu Li}\textsuperscript{3} \quad
\textbf{Xingjian Li}\textsuperscript{5} \\
\textbf{Tianyang Wang}\textsuperscript{1} \quad
\textbf{Jihun Hamm}\textsuperscript{2} \quad
\textbf{Xiao Wang}\textsuperscript{4} \quad
\textbf{Min Xu}\textsuperscript{5}\thanks{Corresponding author: \texttt{mxu1@cs.cmu.edu}} \\[0.5em]
\textsuperscript{1}University of Alabama at Birmingham \quad
\textsuperscript{2}Tulane University \quad
\textsuperscript{3}Brown University \\
\textsuperscript{4}Oak Ridge National Laboratory \quad
\textsuperscript{5}Carnegie Mellon University
}

\begin{document}
\maketitle
\begin{abstract}
Parameter-efficient fine-tuning (PEFT) has emerged as a crucial approach for adapting large vision transformers to downstream tasks without the prohibitive computational costs of full fine-tuning. While existing visual prompt tuning (VPT) methods have made significant strides, they predominantly rely on static, domain-specific prompts that fail to capture the rich visual diversity within individual instances. This paper introduces V$^2$APT (Visual Variational Autoencoder Prompt Tuning), a novel framework that generates dynamic, input-dependent prompts using a variational autoencoder architecture. By learning a latent representation of image-specific features and decoding them into customized prompts, V$^2$APT adapts to the unique visual characteristics of each input. Extensive experiments on FGVC, HTA, and VTAB-1k benchmarks demonstrate that our approach consistently outperforms state-of-the-art PEFT methods. Notably, V$^2$APT achieves +3.2\% improvement over VPT-Deep on HTA, with an average performance gain of +2.0\% across all three datasets.
\end{abstract}    
\section{Introduction}
\label{sec:intro}
\vspace{-1mm}
\textit{``Generative models, including VAEs, enable machines to understand the underlying data distribution, paving the way for creative and unsupervised AI."}  
\vspace{-1.5mm}
\begin{flushright} --- Yann LeCun
\end{flushright}
\vspace{-1.5mm}

Vision Transformers (ViTs)~\cite{dosovitskiy2021image, liu2021swin} have revolutionized computer vision, establishing new performance benchmarks across the visual recognition landscape. Despite their impressive capabilities, harnessing these models for novel applications presents a formidable challenge due to their vast parameter space—often reaching hundreds of millions of trainable weights. Parameter-Efficient Fine-Tuning (PEFT) strategies, particularly Visual Prompt Tuning (VPT)~\cite{jia2022visual}, offer an elegant solution to this computational bottleneck by introducing compact, learnable prompt vectors while preserving the integrity of the pre-trained architecture. This approach has catalyzed numerous innovations, from cross-task cross-domain adaptations~\cite{bahng2022visual,han2023e2vpt,pei2024sa2vp, Zhang_2025_WACV, cai2024bayesian} to multimodal applications~\cite{khattak2023maple,zhou2022unified}.

However, conventional VPT approaches rely on static, randomly initialized prompts—a fundamental limitation that restricts their adaptability \cite{jia2022visual}. This one-size-fits-all strategy overlooks the rich visual diversity within semantic categories, where images sharing the same label (e.g., ``cat") may display vastly different visual characteristics—from close-up portraits to distant silhouettes, from various breeds to different postures. Domain-specific prompts, while efficient, remain blind to these crucial instance variations, resulting in adaptation strategies that cannot fully respond to the unique visual signature of each input. 

To address these limitations, we propose V$^2$APT (Visual Variational Autoencoder Prompt Tuning), a novel framework that dynamically generates input-dependent prompts tailored to individual images. V$^2$APT leverages a Variational Autoencoder (VAE) architecture to capture and encode image-specific features into a latent representation, which is then decoded into customized prompts that adapt to the unique visual properties of each input. This approach enhances both adaptation precision and generalization robustness, enabling more effective transfer learning across diverse visual tasks and domains. 

We validate V$^2$APT on the FGVC \cite{krause20133d}, HTA \cite{huang2023head}, and VTAB-1k \cite{zhai2019large} benchmarks. Our experimental results consistently demonstrate that V$^2$APT outperforms existing state-of-the-art PEFT methods, while maintaining a minimal computational footprint. Specifically, in comparison to VPT-Deep \cite{jia2022visual}, V$^2$APT achieves a substantial absolute improvement of +3.2\% on the HTA benchmark, with even more significant gains of +2.27\% on VTAB-1k. These results highlight generative latent modeling as a promising new paradigm for prompt-based adaptation in ViTs.

\section{Related Works}
\label{sec:related}

\begin{figure*}[t]
    \centering
    \includegraphics[width=0.8\textwidth]{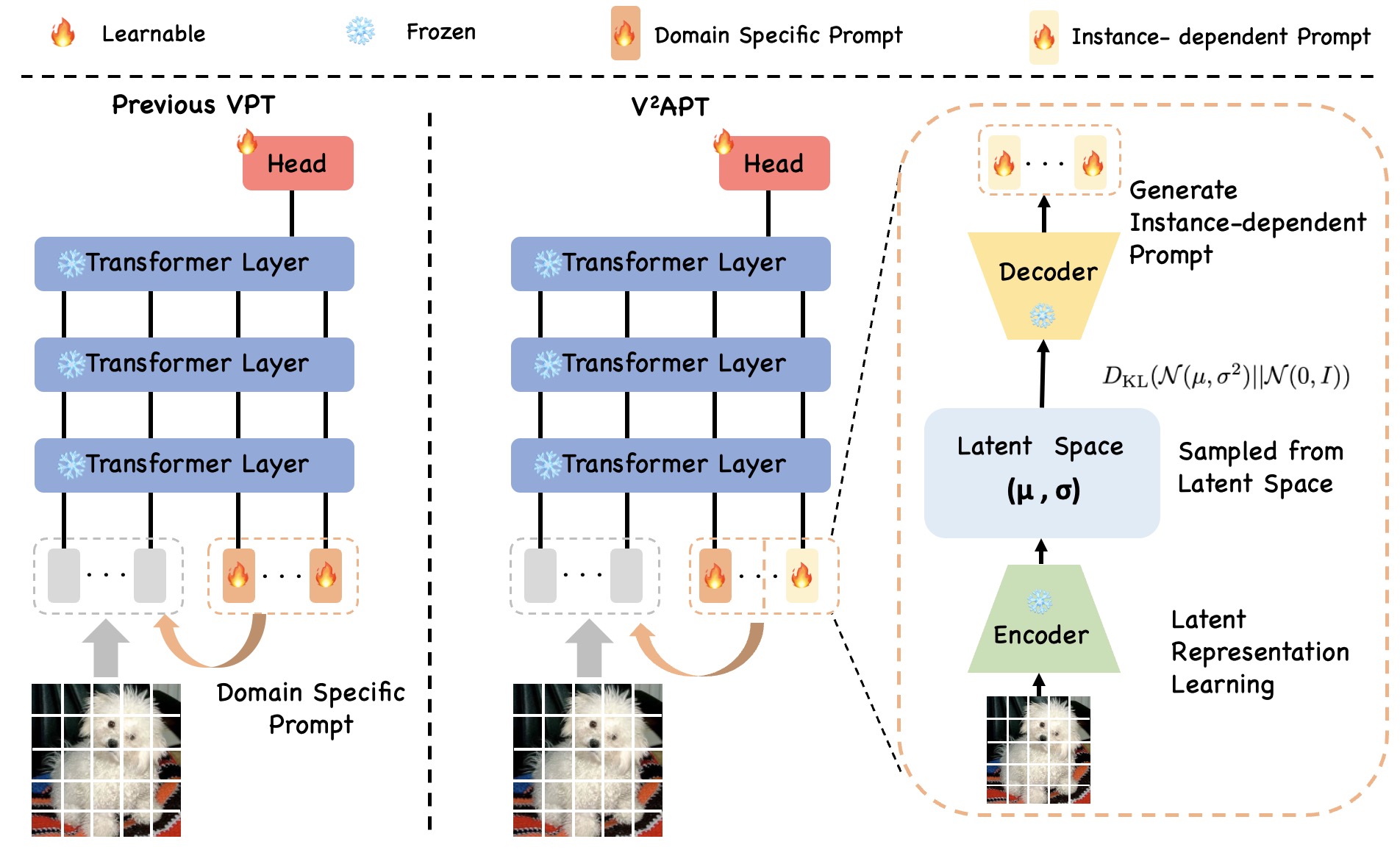}
    \caption{
    Comparison of standard Visual Prompt Tuning (VPT) and our proposed V$^2$APT framework.  
    (\textbf{Left}) VPT employs static \textbf{Domain-Specific Prompts}, which remain fixed across different images.  
    (\textbf{Middle}) Our V$^2$APT framework introduces \textbf{Instance-dependent Prompts}, dynamically generated via a Variational Autoencoder (VAE) based on each input image.  
    (\textbf{Right}) The VAE encodes image embeddings into a KL-divergence-regularized latent space, then decodes them into instance prompts that combine with domain-specific prompts before entering the transformer, maintaining the same token count as VPT.
    }
    \label{fig:framework}
    \vspace{-1mm}
\end{figure*}

\subsection{Visual Prompt Tuning (VPT)}
Visual Prompt Tuning (VPT) enhances Vision Transformers (ViTs) by prepending learnable prompts to inputs \cite{bahng2022visual}, extending to various feature spaces while preserving model architecture and significantly reducing computational overhead \cite{jia2022visual}. Recent advancements have refined this approach through multi-level prompt injection \cite{han2023e2vpt} and spatially aware prompts for improved feature alignment \cite{pei2024sa2vp}. The versatility of VPT has further expanded into multi-modal domains such CLIP-based models \cite{khattak2023maple,zhou2022unified}. Despite these innovations, existing approaches rely exclusively on static soft prompts, overlooking the potential of structured prior knowledge. While numerous VPT variants \cite{chen2022revisiting,cheng2023e2vpt,zeng2024visual,he2023parameter,huang2023hta,huang2023diversity,khattak2023maple,Zhang2024DPCoreDP} have improved performance across diverse tasks, none have explored data-driven prompt generation from latent space. Our work addresses this fundamental gap by integrating VAEs to generate dynamic, structured prompts, effectively bridging the divide between hard and soft prompting paradigms to enhance generalization and robustness in ViTs.

\subsection{VAEs for Representation Learning}
Variational Autoencoders (VAEs) \cite{kingma2013auto} have been extensively employed for learning structured latent representations by enforcing a prior distribution on the latent space. In vision tasks, VAEs effectively capture underlying data distributions, facilitating generative modeling, feature regularization, and domain adaptation \cite{zhao2017learning, tolstikhin2017wasserstein,ma2025cadvaeleveragingcorrelationawarelatents,higgins2017beta}. While recent research has begun exploring VAEs for adaptive representations in transformers \cite{xu2022variational}, their potential for prompt learning remains largely unexplored. Our work bridges this gap by introducing VAE-based instance-specific image prompts, where the regularized latent space models semantic structure across diverse visual inputs. This approach ensures adaptive and task-specific tuning that responds to individual image characteristics while maintaining computational efficiency.
\section{Methodology}
\label{sec:methodology}

In this section, we present V$^2$APT, a novel visual prompt tuning framework that integrates Variational Autoencoders (VAEs) to dynamically generate instance-specific prompts. By leveraging latent distribution modeling, V$^2$APT adapts prompts to the underlying data structure, improving task relevance and adaptability. We first introduce the fundamentals of visual prompt tuning in \S~\ref{sec:preliminary}, followed by the detailed description of V$^2$APT in \S~\ref{sec:vae-prompt}. The overall architecture is illustrated in Fig.~\ref{fig:framework}.

\subsection{Preliminary}
\label{sec:preliminary}

\textbf{Visual Prompt Tuning.} Visual Prompt Tuning (VPT) introduces a set of learnable prompts $\mathbf{P} = \{\mathbf{P}^1, \mathbf{P}^2, \ldots, \mathbf{P}^N\}$ into a pre-trained ViT model $\mathcal{T}$ with $N$ transformer layers. These $d$-dimensional prompts are prepended to the input embeddings and optimized during training while keeping the rest of the model frozen. For the first layer, the contextual embeddings are computed as $\mathbf{Z}^1 = L_1(\mathbf{P}^1, \mathbf{E})$, where $\mathbf{E}$ denotes the frozen patch embeddings of input image patches, and $\mathbf{P}^1$ represents the trainable prompts. For subsequent layers, the embeddings are iteratively updated using $\mathbf{Z}^i = L_i(\mathbf{P}^i, \mathbf{Z}^{i-1})$. However, these prompts are domain-specific, meaning they remain fixed for a given dataset and do not adapt to variations within individual samples. This static nature limits their effectiveness in handling diverse input distributions or task-specific characteristics.

\subsection{Visual Variational Autoencoder Prompt Tuning}
\label{sec:vae-prompt}

To address the limitations of domain-specific prompts, V$^2$APT introduces a Variational Autoencoder (VAE)-based mechanism for generating adaptive prompts that dynamically align with the latent structure of each input image.

\noindent\textbf{Instance-dependent Prompt Generation.} In V$^2$APT, the VAE encoder maps the input embeddings $\mathbf{X}$ into a latent distribution $\mathcal{N}(\mu, \sigma^2)$, where $\mu$ and $\sigma$ represent the mean and variance of the latent variables, respectively, i.e., $\mathbf{Z} \sim \mathcal{N}(\mu, \sigma^2) = \text{Encoder}(\mathbf{X})$. The sampled latent variables $\mathbf{Z}$ are then passed through a decoder to generate instance-specific image prompts, formulated as $\mathbf{P}^{\text{inst}} = \text{Decoder}(\mathbf{Z})$. These prompts $\mathbf{P}^{\text{inst}} = \{\mathbf{P}_1, \mathbf{P}_2, \ldots, \mathbf{P}_N\}$ are then injected into the Transformer encoder layers. Unlike domain-specific prompts, instance-dependent prompts are dynamically generated for each input image, improving task relevance and representation learning.

\noindent\textbf{KL Divergence Regularization.} To ensure that the latent variables $\mathbf{Z}$ align with a standard Gaussian prior $\mathcal{N}(\mathbf{0}, \mathbf{I})$, we incorporate a KL divergence regularization term:
\begin{equation}
    \mathcal{L}_{\text{KL}} = D_{\text{KL}} \big( \mathcal{N}(\mu, \sigma^2) \,||\, \mathcal{N}(\mathbf{0}, \mathbf{I}) \big).
\end{equation}
This constraint regularizes the latent space, enhancing generalization and preventing overfitting.

\noindent \textbf{Final Prompt Composition.} The final prompt input to each Transformer encoder layer is a concatenation of domain-specific and instance-dependent prompts, defined as $\hat{\mathbf{P}}_i = [\mathbf{P}^{\text{inst}}_i, \mathbf{P}^{\text{dom}}_i]$, where $\mathbf{P}^{\text{inst}}_i$ represents the VAE-generated instance-dependent prompts, and $\mathbf{P}^{\text{dom}}_i$ corresponds to the static domain-specific prompts. Importantly, the total number of prompt tokens remains the same as in standard VPT, ensuring that V$^2$APT does not introduce additional computational overhead.
\section{Experiment}
\label{sec:experiments}

\begin{table*}[!htbp]
\caption{\textbf{Performance comparison on the FGVC benchmark with ViT.} }
\vspace{-2mm}
\centering
\setlength{\tabcolsep}{0.8pt} 
\small 
\begin{tabular*}{\textwidth}{@{\extracolsep{\fill}}lccccccc@{}}
\toprule
\textbf{Methods} & \textbf{CUB-200-2011} & \textbf{NABirds} & \textbf{Oxford Flowers} & \textbf{Stanford Dogs} & \textbf{Stanford Cars} & \textbf{Mean} \\
\midrule
Full Fine-tune\cite{iofinova2022well} & 87.3 & 82.7 & 98.8 & 89.4 & \textbf{84.5} & 88.5  \\
Head Fine-tune\cite{iofinova2022well} & 85.3 & 75.9 & 97.9 & 86.2 & 51.3 & 79.3  \\
VPT-deep\cite{jia2022visual} & 88.5 & 84.2 & \textbf{99.0} & 90.2 & 83.6 & 89.1  \\
\textbf{Ours} & \textbf{89.1} & {\textbf{85.0}} & {\textbf{99.0}} & \textbf{90.9} & 83.8 & \textbf{89.6} \\
\bottomrule
\end{tabular*}
\label{tab:fgvc}
\end{table*}

\begin{table*}[!htbp]
\centering
\caption{\textbf{Performance comparison on the HTA benchmark with ViT.} }
\vspace{-2mm}
\setlength{\tabcolsep}{0.9pt} 
\small 
\begin{tabular*}{\textwidth}{@{\extracolsep{\fill}}lccccccccccc@{}}
\toprule
\textbf{Methods} & \textbf{DTD} & \textbf{CUB-200} & \textbf{NABirds} & \textbf{Dogs} & \textbf{Flowers} & \textbf{Food-101} & \textbf{CIFAR-100} & \textbf{CIFAR-10} & \textbf{GTSRB} & \textbf{SVHN} & \textbf{Mean} \\
\midrule
Full Fine-tune\cite{iofinova2022well} & 64.3 & 87.3 & 82.7 & 89.4 & 98.8 & 84.9 & 68.9 & 97.4 & {\textbf{97.1}} & 87.4 & 85.8 \\
Head Fine-tune\cite{iofinova2022well} & 63.2 & 85.3 & 75.9 & 86.2 & 97.9 & 84.4 & 63.4 & 96.3 & 68.0 & 36.6 & 75.7 \\
VPT-deep\cite{jia2022visual} & 65.8 & 88.5 & 84.2 & 90.2 & \textbf{99.0} & 83.3 & 78.8 & 96.8 & 90.7 & 78.1 & 85.5 \\
\textbf{Ours} & \textbf{68.1} & \textbf{89.1} & {\textbf{85.0}} & \textbf{90.9} & \textbf{99.0} & \textbf{88.6} & \textbf{82.8} & \textbf{97.6} & 96.2 & \textbf{90.0} & \textbf{88.7} \\
\bottomrule
\end{tabular*}
\label{tab:hta}
\end{table*}

\subsection{Experimental Setup}

\noindent \textbf{Datasets.} 
We evaluate our method on three comprehensive benchmarks: FGVC, HTA, and VTAB-1k \cite{zhai2019vtab} to assess adaptability and robustness across diverse visual tasks. 
The \textbf{FGVC} benchmark comprises five fine-grained datasets (CUB, NABirds, Oxford Flowers, Stanford Dogs, and Stanford Cars) designed to evaluate discrimination between subtle category variations, following the splits established in \cite{jia2022visual}. 
The \textbf{HTA} benchmark covers 10 datasets including CIFAR10, CIFAR100, DTD, CUB-200, NABirds, Oxford Flowers, Food101, GTSRB, and SVHN, testing generalization capabilities across domains with the configuration from DAM-VP \cite{huang2023hta}. 
The \textbf{VTAB-1k} benchmark spans 19 datasets categorized into `Natural,' `Specialized,' and `Structured' groups, each containing 1000 images (800 for training, 200 for validation) to thoroughly evaluate model robustness under limited data conditions.


\noindent \textbf{Implementation Details.} 
We employ ViT-B/16 and Swin Transformer architectures, both pre-trained on ImageNet-21K \cite{deng2009imagenet, liu2021swin}. 
During fine-tuning, we keep the backbone frozen, updating only the prompt and classification head parameters. 
We optimize using AdamW \cite{loshchilov2017adamw} with a learning rate of $1 \times 10^{-3}$, weight decay of $1 \times 10^{-4}$, and batch sizes of 64 or 128 depending on the dataset. 
Classification accuracy serves as our primary evaluation metric across all experiments \cite{pei2024sa2vp}.

\subsection{Experimental Results}

\noindent \textbf{Performance on FGVC and HTA benchmarks with ViT Backbone.}  
V$^2$APT consistently outperforms baselines across various fine-grained and general visual benchmarks, as shown in Table~\ref{tab:fgvc} and Table~\ref{tab:hta}. On FGVC datasets, our method achieves state-of-the-art accuracy, particularly excelling in animal classification tasks, where VAE-driven prompts enhance feature adaptation by capturing task-relevant variations. The latent space modeling enables more adaptive prompt generation, improving generalization. Notably, on Stanford Cars, while Full Fine-tune benefits from full adaptation, V$^2$APT surpasses all parameter-efficient methods, demonstrating its effectiveness in structured fine-grained recognition. On the HTA benchmark, V$^2$APT maintains strong performance across diverse datasets. It achieves the highest accuracy on DTD (texture) and Food-101 (food classification), showcasing its ability to adapt to domain-specific variations. Moreover, on lower-resolution datasets such as CIFAR-10, CIFAR-100, and SVHN, our method remains in the top-3, demonstrating robustness to varying image quality and scale. These results confirm that V$^2$APT effectively generalizes across multiple visual recognition tasks, outperforming SOTA PEFT methods.


\renewcommand{\floatpagefraction}{0.9} 
\renewcommand{\textfraction}{0.1} 


\begin{figure}[t]
    \centering
    \subfloat[ViT]{
      \begin{minipage}[b]{0.8\linewidth} 
        \centering  
        \includegraphics[width=\linewidth]{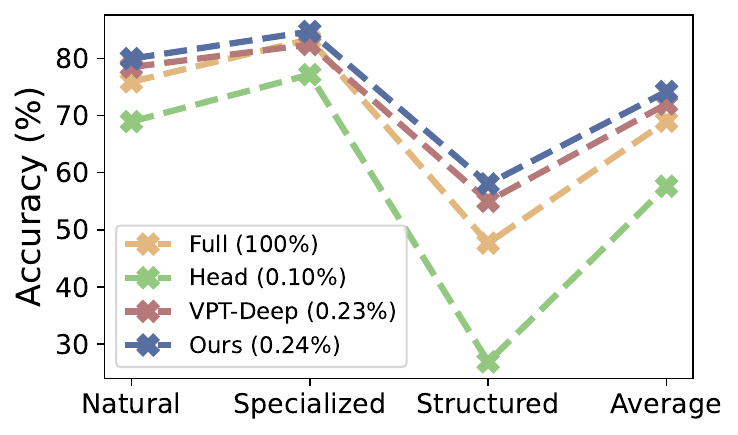} 
        \vspace{-3mm}
        \label{fig:vit_res}
      \end{minipage}
    }
    \\
    \subfloat[Swin Transformer]{
      \begin{minipage}[b]{0.8\linewidth} 
        \centering  
        \includegraphics[width=\linewidth]{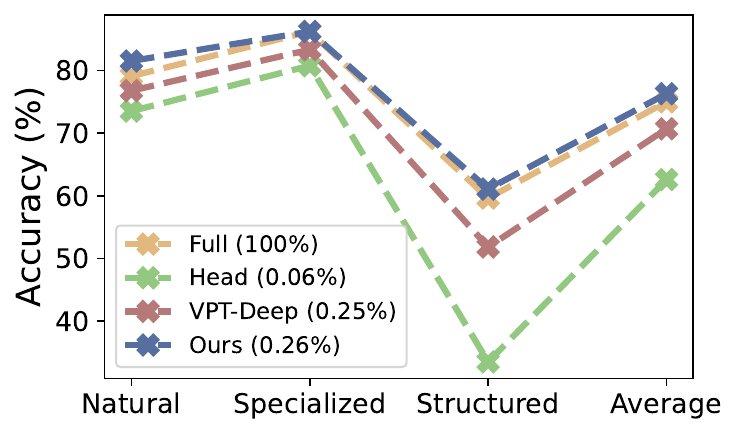} 
        \vspace{-3mm}
        \label{fig:vit_swin}
      \end{minipage}%
    }
    \caption{
    Comparison of fine-tuning methods on VTAB-1k across different models. The number within parentheses indicates the percentage of trainable parameters for each approach. Our method consistently outperforms existing techniques on both ViT and Swin transformer architectures.
}
    \vspace{-2mm}
\end{figure}

\noindent \textbf{Performance on VTAB-1k with ViT and Swin Transformer.}  
Fig.~\ref{fig:vit_res} and Fig.~\ref{fig:vit_swin} show that V$^2$APT consistently achieves the highest accuracy across all three VTAB-1k categories, demonstrating superior adaptability. This improvement stems from the learned latent representations, which enhance adaptation to domain-specific variations. The largest gain is observed in Structured tasks, which require spatial reasoning. Our method effectively captures both global and local patterns through task-aware feature modeling, leading to better representation learning. Notably, V$^2$APT maintains its advantage across different backbone architectures, further validating its robustness and generalization capabilities. These results confirm that our approach outperforms static prompt tuning, providing more adaptive and context-aware fine-tuning.

\begin{figure}[t]
    \centering
    \includegraphics[width=0.75\linewidth]{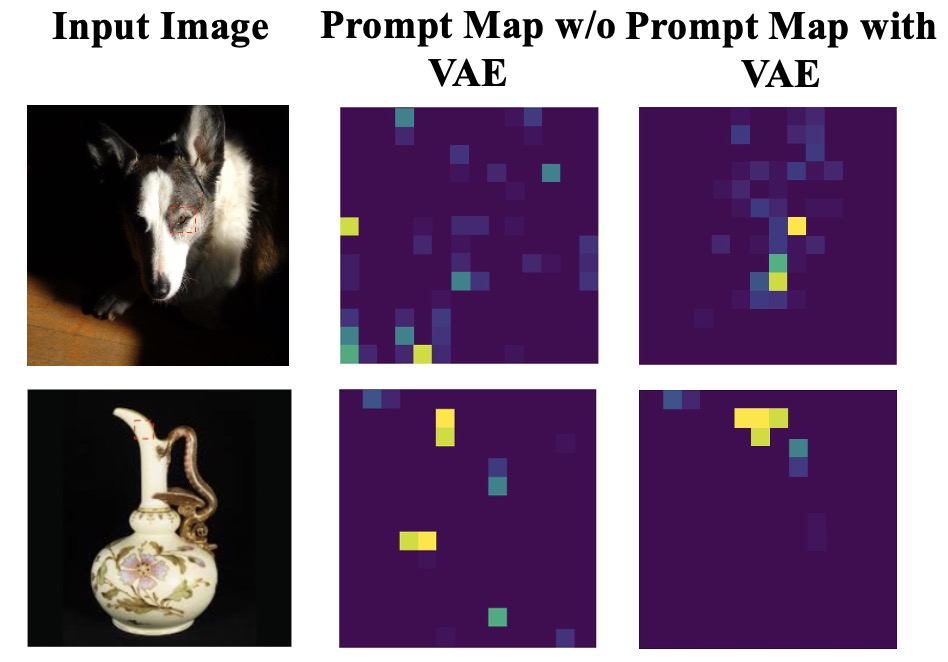}
    \caption{Visualization of prompt maps with and without VAE. The first column shows input images, while the second and third columns illustrate the learned prompt maps without and with VAE integration, respectively.}
    \label{fig:prompt_map}
\end{figure}

\noindent \textbf{Cosine Similarity Analysis.} 
We firstly investigate whether the learned features well match image clues. To this end, we adopt the cosine similarity map \cite{vaswani2017attention} that was developed for a similar purpose. As illustrated in Fig. \ref{fig:prompt_map}, without the aid of our instance image  prompt, the similarity between the learned prompt and features is much lower (i.e, \textbf{middle}), indicating a higher degree of mismatch, while the opposite is observed (i.e., \textbf{right}) when instance image prompt is leveraged, indicating that such prompt benefits feature learning. 
\section{Conclusion}
\label{sec:conclusion}

We introduced V$^2$APT, a novel framework that generates dynamic, input-dependent prompts for Vision Transformers using variational autoencoding. By responding to the unique visual characteristics of individual inputs rather than relying on static prompts, our approach consistently outperforms state-of-the-art PEFT methods across FGVC, HTA, and VTAB-1k benchmarks, particularly excelling on structured tasks with complex domain shifts. Notably, V$^2$APT achieves an improvement of +2.0\% over VPT-Deep across three benchmarks while using a comparable parameter budget. V$^2$APT bridges the gap between soft and hard prompting paradigms while maintaining minimal computational overhead, establishing a promising direction for enhancing both adaptation and generalization in vision models.
\clearpage
{
    \small
    \bibliographystyle{ieeenat_fullname}
    \bibliography{main}
}


\end{document}